\documentclass{Interspeech2024}
\usepackage{authblk}  

\interspeechcameraready

\title{Exploring Energy-Based Models for Out-of-Distribution Detection in Dialect Identification}
\interspeechcameraready

\name{Yaqian}{Hao$^\dagger$}
\name{Chenguang}{Hu$^\dagger$}
\name{Yingying}{Gao}
\name{Shilei}{Zhang$^*$}
\name{Junlan}{Feng$^*$}

\address{
  China Mobile Research Institute, Beijing, China}
 \email{\{haoyaqian, huchenguang,  gaoyingying, zhangshilei, fengjunlan\}@chinamobile.com}

\keywords{Energy-based models, out of distribution detection, dialect identification,  open set classification
}

\begin{document}

\maketitle

\begin{abstract}
    
The diverse nature of dialects presents challenges for models trained on specific linguistic patterns, rendering them susceptible to errors when confronted with unseen or out-of-distribution (OOD) data. This study introduces a novel margin-enhanced joint energy model (MEJEM) tailored specifically for OOD detection in dialects. By integrating a generative model and the energy margin loss, our approach aims to enhance the robustness of dialect identification systems.
Furthermore, we explore two OOD scores for OOD dialect detection, and our findings conclusively demonstrate that the energy score outperforms the softmax score. Leveraging Sharpness-Aware Minimization to optimize the training process of the joint model, we enhance model generalization by minimizing both loss and sharpness.
Experiments conducted on dialect identification tasks validate the efficacy of Energy-Based Models  and provide valuable insights into their performance.
\end{abstract}

\renewcommand{\thefootnote}{\fnsymbol{footnote}}
\footnote{Corresponding Author.}
\footnote{Equal Contribution.}
\renewcommand{\thefootnote}{\arabic{footnote}}
\section{Introduction}
\label{into}
Speech dialect is fundamental in various real-world scenarios, such as enhancing automatic speech recognition systems and facilitating language translation services \cite{zhang2018language,das2021multi}. The accurate identification of spoken dialects  is imperative for delivering tailored and contextually appropriate services. Nevertheless, practical applications often encounter unknown dialects and languages, presenting a significant challenge.   Therefore,
models employed for dialect identification require the capability to discern out-set data, ensuring robustness and reliability, and preventing inaccurate predictions for unseen dialects across diverse linguistic contexts \cite{rakib2023ood,das2023unsupervised,zheng2020out}.
One approach to OOD detection employs the classifier-based method, which distinguishes OOD samples by analyzing the prediction outputs of the logits. However, the softmax confidence score may encounter challenges in reliably distinguishing between in-distribution (ID) and OOD data \cite{liu2020energy,lin2021mood}. Another OOD detection strategy is the density-based method, which estimates the likelihood score $\log p(x)$ of the training data and rejects samples with low likelihood scores. Nonetheless, this method is susceptible to issues associated with overestimating data density. Energy-based models (EBMs), a promising class of density estimators known for their unrestricted architecture, have demonstrated potential in OOD detection \cite{lecun2006tutorial,song2021train}. Additionally, EBMs and softmax-based classifiers are related through the utilization of the softmax function, which transforms energy functions into probability distributions \cite{liu2020energy,grathwohl2019your}. Consequently, leveraging EBMs enables simultaneous exploration of classifier-based and density-based OOD detection methodologies.

In this study, we first propose a margin-enhanced joint energy model 
 incorporating a discriminative classification model, energy-based margin loss and a generative model for dialect identification.  We investigate the effectiveness of energy functions in outlier detection, and  we further assess the OOD detection performance by leveraging energy score and softmax score. 
We conduct numerical experiments on dialect OOD detection task, optimizing model training with the Sharpness-Aware Minimization (SAM) approach. The results demonstrate the effectiveness of the proposed energy model, and show that energy score outperforms traditional softmax score.
Additionally, the ablation  study investigates the impact of margin loss and SAM methods to the performance of dialect OOD detection. Our findings indicate 
despite the inclusion of auxiliary datasets in OE, EBMs achieves comparable  performance, hinting at their innate advantage in dialect OOD detection.\\
Our contributions can be summarized as follows:
\begin{enumerate}
    \item We propose a margin-enhanced energy joint 
 model (MEJEM) for dialect OOD detection that integrates an energy regularization loss and a generative model. Furthermore, we apply the SAM method to improve the optimization of proposed joint model. 
    \item We introduce a variety of EBMs for dialect OOD detection and conduct comprehensive experiments to analyze their performance in dialect OOD detection tasks, demonstrating their effectiveness.
    \item We further investigate the efficacy of energy score and softmax score of different models on dialect OOD detection, finding that the energy score yields superior performance.
\end{enumerate}

\section{Related Work}
\textbf{Classifier-based and Density-based OOD detection.}
The classifier-based OOD detection methods, encompasses various techniques such as the maximum softmax score, ODIN, maximum logits score, and maximum entropy predictions \cite{hendrycks2016baseline,hendrycks2019scaling}. \cite{hsu2020generalized} refined this methodology by employing temperature scaling techniques.
These methods leverage the output probabilities of the neural network classifier to discern between ID and OOD samples. In contrast, Density-based OOD detection methods focus on modeling the underlying data distribution \cite{zhai2016deep,elflein2023out} or estimating the distribution over activations at multiple layers of the neural network \cite{zisselman2020deep,nalisnick2018deep}. 
These techniques, employing density estimation or generative modeling, aim to detect out-set samples by comparing the distributions of ID and OOD data.
\\
\textbf{OOD detection with EBMs.} 
As efficient training methods for deep EBMs have advanced, there has been a growing interest in utilizing EBMs for detecting OOD data.
\cite{liu2020energy} introduced energy scores, showing their superiority over softmax scores, especially in image datasets. Joint EBMs, proposed by \cite{grathwohl2019your}, have shown effectiveness in OOD detection tasks,  open-world EBMs were introduced by \cite{wang2021energy} subsequently. Further efforts include optimizing EBM training to mitigate OOD data density overestimation \cite{kim2022mitigating}, and developing energy-based meta-learning frameworks for robust OOD task generalization \cite{chen2024secure}.

However, research on speech dialect OOD detection is relatively limited \cite{das2023unsupervised}, and there exists no previous work that has examined the application of EBMs for this particular task.
In this study, we present MEJEM framework combining energy-based margin loss  with joint energy model and assess the efficacy of EBMs within the domain of  dialect identification.


\section{Method}
\label{section:method}
\subsection{Energy-based Models}
EBMs are  probabilistic models defined by energy function \(E(\mathbf{x}; \theta)\), where \(\mathbf{x}\) represents the input data and \(\theta\) denotes the model parameters \cite{lecun2006tutorial,ou2024energy}. The  probability distribution derived from the Boltzmann distribution, is given by:
\begin{equation}
 p(\mathbf{x}; \theta) = \frac{1}{Z(\theta)} \exp(-E(\mathbf{x}; \theta)), 
 \end{equation}
where $Z(\theta) = \int_{\mathbf{x}} \exp(-E(\mathbf{x}; \theta)) d\mathbf{x} $, is the partition function.\\
 For a discriminative classifier, which map input data \(\mathbf{x}\) to class logits \(f_\theta(\mathbf{x})\) using the softmax function, the probability of class \(y\) given input \(\mathbf{x}\) is calculated as:
\begin{align} \label{eq:py}
p_\theta(y | \mathbf{x}) = \frac{\exp(f_\theta(\mathbf{x})[y])}{\sum_i^K \exp(f_\theta(\mathbf{x})[i])}, \end{align}
where \( K \) is the number of classes.
This mapping allows us to interpret the classifier \(f(\mathbf{x})\) as an energy function \(E_\theta(\mathbf{x}, y) = -f_\theta(\mathbf{x})[y]\) within the EBM framework~\cite{grathwohl2019your,liu2020energy}. 
Furthermore, the logits \(f_\theta(\mathbf{x})\) facilitate the formulation of an EBM for the joint distribution of data point $(\mathbf{x},y)$:      $p_\theta(\mathbf{x}, y) =\exp(f_\theta(\mathbf{x})[y])/Z(\theta)$.
Marginalizing over \(y\) yields an unnormalized density model for \(\mathbf{x}\):
\begin{align}\label{eq:px} p_\theta(\mathbf{x}) = \sum_y p_\theta(\mathbf{x}, y) = \sum_y \frac{\exp(f_\theta(\mathbf{x})[y])}{Z(\theta)}, \end{align} 
thus defining an EBM. 
\subsection{Joint Margin-based Energy Model}
\textbf{Motivation.}
The perspective illustrated in Eq.~\eqref{eq:py} to Eq.~\eqref{eq:px} underscores the inherent compatibility between the softmax classifier and the EBM, facilitating a unified understanding of their shared principles. It has been demonstrated that the joint EBM (JEM) framework could significantly enhance OOD detection performance for image classification while preserving classification precision \cite{grathwohl2019your}. The log-likelihood \(p_\theta(x,y)\) in JEM serves as a unified objective function, encompassing both classification loss and the likelihood score derived from the generative model, thereby providing a comprehensive evaluation of ID data.
However, the energy function \(E(\mathbf{x}; \theta)\) solely focuses on capturing the density of  training data, thus lacking knowledge of the boundary between ID data and OOD data.   To address this limitation, leveraging the energy function with auxiliary outlier data can enhance the model’s capability to identify OOD.
\\
\textbf{Margin-Enhanced Joint Energy  Model (MEJEM).}
Inspired by the motivation above, we propose a margin-enhanced joint energy model, which integrates a discriminative model, a generative model, and energy-based margin loss. The objective function of MEJEM is:
\begin{equation} \label{eq-joint}
   L(\theta)=  \log p_\theta(y | x) + \lambda_1\log p_\theta(x) +\lambda_2 L_{e},
\end{equation} 
where \(\lambda_1\) and \(\lambda_2\) are hyperparameters balancing the cross-entropy loss, generative loss, and energy-based margin loss. \\
\textbf{Hybrid Energy-based Margin Loss.}
Outlier data training may lead to a broader distribution of errors due to the presence of anomalies or noise in OOD data. To address this issue,
we present a hybrid energy-based margin loss function that combines hinge loss and square loss, incorporating auxiliary outlier data during training. For each sample \(x_i\), the hybrid energy loss is defined as follows:
\begin{equation}
L_{e}(x_i) =
\begin{cases} 
\left(\max(E_{x_i} - M_{in},0)\right)^2,& \text{if } (x_i,y_i )\in  D_{in}, \\
\max( M_{out}-E_{x_i},0),& \text{if } (x_i,y_i) \in D_{out},
\end{cases}
\end{equation}
where $D_{in}$ and $D_{out}$ denote the ID and OOD training samples, respectively, with $K$ classes for ID training data. We utilize linear hinge loss for OOD data and square hinge loss for ID data, achieving a balanced approach with moderate error penalties to mitigate overfitting risks.  
Thus, the total energy-based margin loss is given as following:
\begin{equation}
   L_e= \sum_{ D_{in}\cup D_{out}} L_{e}(x_i).
\end{equation}

\subsection{Classifier-based and Energy-based OOD}
Classifier-based OOD detection methods rely on discriminative neural classifiers to distinguish between ID and OOD samples. Initially, class probabilities \( P(y | \mathbf{x}) \) are computed for an input sample \( \mathbf{x} \) using the softmax function in Eq.~\eqref{eq:py}.
Subsequently, the OOD score is determined as the maximum of class probabilities:
\begin{equation}
Score_{c} = \max_{y} P(y | \mathbf{x}).
\end{equation}
The energy-based  OOD score can be interpreted from a density perspective by utilizing the energy function, which represents the logarithm of the data density \( p(\mathbf{x}) \):
\begin{equation} Score_e = -E_\theta(\mathbf{x}; f) = \log \sum_i^K \exp(f_\theta(\mathbf{x})[i]). 
\end{equation}
In both classifier-based and energy-based OOD detection methods,  
the label \( \hat{y} \) is predicted using a piecewise function based on the OOD score as follows:
\begin{equation}
\hat{y} = 
\begin{cases} 
K + 1 & \text{if } \text{OOD Score} < \delta, \\
\text{argmax} \, P(y | \mathbf{x}) & \text{otherwise}.
\end{cases}
\end{equation}
If OOD score is below the threshold, \( \hat{y} \) becomes \( K + 1 \); otherwise, it belongs to the class with the highest probability \( P(y | \mathbf{x}) \).
\vspace{-0.4cm}
\begin{figure*}[htbp]
\centering 
\includegraphics[width=0.910\linewidth,height=0.31\linewidth]{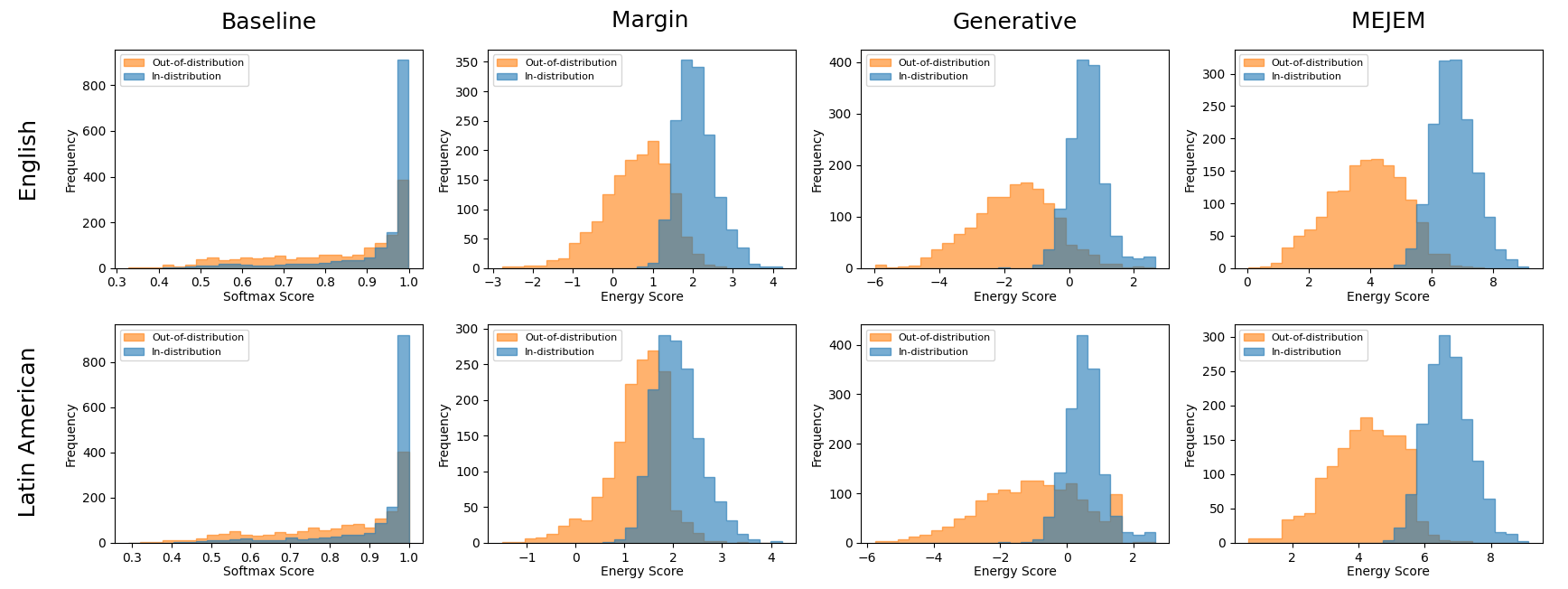} \vspace{-0.331cm}
\caption{Comparison of OOD distribution.} 
\label{distribution} 
\end{figure*}
\subsection{Sharpness-aware minimization}
To enhance the optimization of the joint objective function Eq.~\eqref{eq-joint}, we adopt Sharpness-Aware Minimization (SAM), a method that concurrently minimizes loss value and sharpness by identifying parameters within neighborhoods of uniformly low loss values \cite{foret2020sharpness}. This approach can help prevent EBMs from converging to excessively sharp local optima \cite{foret2020sharpness,yang2023towards}. By integrating SAM methods into MEJEM, the minimax objective function defined in Eq.~\eqref{eq-joint} can be formulated as follows:
\begin{equation}
\max_{\theta} \min_{\Vert \epsilon \Vert_2 \leq \rho  } L(\theta+\epsilon) +\beta \Vert \theta \Vert_2^2,
\end{equation}
where $\rho$ denotes the radius of the $L_2$ norm ball centered at the model parameters $\theta$, serving as a constraint on their deviation, while $\beta$ represents a hyperparameter that tunes the intensity of $L_2$ regularization imposed upon $\theta$.
\subsection{Stochastic Gradient Langevin Dynamics}
For the outer maximization of data distribution $p_\theta(x)$, we utilize SGLD to sample from $p_\theta(\mathbf{x})$. The process starts with sampling $x_0$ from a simple prior distribution and running an overdamped Langevin diffusion for $K$ steps with a positive step size $\epsilon > 0$. The update for each step $k = 0, 1, \ldots, K - 1$ is as follows:
\begin{equation}
    x_{k+1} = x_k - \frac
    {\epsilon^2}{2}  \nabla_{x_k} E_\theta(x_k)  + \epsilon z_k,
\end{equation}
where $\nabla_{x_k} E_\theta(x_k)$ is the gradient of the energy function with respect to $x_k$, and $z_k$ is a random noise term. As $\epsilon \rightarrow 0$ and $K \rightarrow \infty$, the final sample $x_K$ converges to $p_\theta(x)$ under certain regularity conditions. 

\section{Experiments}
\label{section:exp}
\subsection{Datasets}
We have used 4 open-source datasets OLR2021 (Oriental Language Recognition Challenge 2021)~\cite{wang2021olr}, UK and Ireland English Dialect speech dataset~\cite{demirsahin2020open}, Latin American Spanish corpus~\cite{guevara2020crowdsourcing} and Common Voice~\cite{ardila2019common}. OLR2021 includes 5 Chinese dialects which are used as fixed known classes for training. OOD samples are randomly selected from English dialect (Welsh, Scottish, Irish, Northern, Midlands, Southern) from~\cite{demirsahin2020open} and Spanish dialect (Argentinian, Chilean, Colombian, Peruvian, Venezuelan and Puerto Rican) from~\cite{guevara2020crowdsourcing} respectively for evaluation.
We also select samples from~\cite{ardila2019common} as auxiliary OOD data for outlier exposure training. 
More details on the respective train and test set are provided in Table~\ref{data}.

\subsection{Experimental Settings}

All our experiments are based on the Wide-ResNet architecture~\cite{zagoruyko2016wide}, featuring a width of 3, depth of 22,  SGLD ~\cite{welling2011bayesian} learning rate of 0.1, 15 sampling steps, batch size of 128, and a buffer size of 10,000, with training conducted using the PyTorch toolkit.
We optimize the model with stochastic gradient descent~\cite{bottou2010large} optimizer, the learning rate warms up to 0.1 during the first 1000 steps, and is reduced at epoch [35, 70, 100] with a decay rate of 0.2. For the speech feature, a 192-frame segment is randomly chunked from each utterance. The input features are 32-dimensional Mel Filter-Banks (Fbanks)
extracted using the librosa package~\cite{mcfee2015librosa} 
with a window length of 25ms and a shift of 10ms with Hamming window. Mean and variance normalization is applied to Fbanks. 
We also explored using SpecAugment~\cite{park2019specaugment}, Noise and Reverberation for data augmentation to improve classification precision.
However, this introduced significant instabilities during generative training, leading to early collapse.
\vspace{-0.23cm}
\begin{table}[htbp]
\centering
\caption{Dataset}\label{data}\vspace{-0.002cm}
\label{tab:data}
\small\resizebox{6.2825cm}{!}{%
\begin{tabular}{ccccc} 
\hline
Corpus      & Class        & Spk & Train\_utt & Test\_utt  \\ 
\hline

            & Shanghainese   & 21  & 3000       & 300       \\
            & Sichuanese   & 21  & 3000       & 300       \\
OLR2021 & Hokkien   & 27  & 3000       & 300       \\
            & Mandarin   & 24  & 3000       & 300       \\
            & Cantonese  & 24  & 3000       & 300       \\
\hline
Common Voice    & -  & 1000  & 31775       & -       \\
\hline
Latin American& 6 Dialects & 174&- & 1500 \\
English& 6 Dialects & 120&- & 1500 \\

\hline
\end{tabular}}
\end{table}
\vspace{-0.61cm}
 \begin{table}[htbp]
\centering
\small
\caption{Close-set Performance}
\label{tab:close}
 \renewcommand{\arraystretch}{0.9670} 
\resizebox{6.538cm}{!}{
\begin{tabular}{ccc}
\hline
Model  & Precision(\%)$\uparrow$ & Relative Change(\%)\\
\hline
WideResNet    & 99.23 & - - \\
    WideResnet(SAM)   &99.36 &+0.13\%($\uparrow$)  \\
OE(SAM)~\cite{hendrycks2018deep}  &99.36 &+0.13\%($\uparrow$)\\
Energy Margin   & 98.91&-0.32\%($\downarrow$) \\
SADAJEM~\cite{yang2023towards}  & \textbf{99.55} &+0.32\%($\uparrow$)  \\
JEMPP~\cite{yang2021jem++}   & 99.08 &-0.15\%($\downarrow$)  \\MEJEM  & 98.64 &-0.59\%($\downarrow$)\\
\hline
\end{tabular}
}
\end{table}
\vspace{-0.615cm}\\
\section{Results and Discussion}
\begin{table*}[htbp]
\centering
\small
\caption{OOD task Detection Performance}
\label{tab:ood-performance}
 \renewcommand{\arraystretch}{0.937} 
\resizebox{0.889\textwidth}{!}{
\begin{tabular}{ccccccccccc}
\hline
&Model & Generative & Margin & SAM & Auxiliary  &Score & \multicolumn{2}{c}{Latin American} & \multicolumn{2}{c}{English}  \\
\cline{8-9} \cline{9-11} 
&& Model & Loss &Method &OOD data &Function &  AUROC$\uparrow$ & FPR95$\downarrow$  &AUROC$\uparrow$ & FPR95$\downarrow$ \\
\hline
 &WideResNet~\cite{zagoruyko2016wide} &  &  &&& softmax & \textbf{0.763} & 0.829 & \textbf{0.733} & 0.852  \\
&&  &  &&& energy & 0.684 & 0.812 & 0.589 & 0.936 \\
Softmax &WideResNet+SAM &  &  &$\checkmark$  && softmax & \textbf{0.783} & 0.768 & \textbf{0.739} & 0.808  \\
Model&& & &  && energy &0.729 & 0.761 & 0.609 & 0.875 \\
&OE~\cite{hendrycks2018deep} &  &  & $\checkmark$ &$\checkmark$& softmax&\textbf{0.812} & 0.731  &  \textbf{0.795} & 0.779 \\
&& & &  && energy & 0.763 & 0.716 & 0.772 & 0.779 \\ \hline
&JEMPP & $\checkmark$ &  &  && softmax & 0.595 & 0.905 & 0.527 & 0.929 \\
& & & &  &&energy & \textbf{0.770} & 0.589 & \textbf{0.836} & 0.401  \\
 Energy&SADAJEM & $\checkmark$ &  &$\checkmark$ && softmax & 0.752 & 0.865 & 0.729 & 0.874  \\
 Model  && &  &  && energy & \textbf{0.829} & 0.340 & \textbf{0.967} & 0.126 \\
&MEJEM & $\checkmark$ & $\checkmark$ & $\checkmark$ &$\checkmark$& softmax & 0.784 & 0.784 & 0.774 & 0.758 \\
&& & &  && energy & $\textbf{0.925}^*$ & $\textbf{0.223}^*$  & $\textbf{0.987}^*$ & $\textbf{0.054}^*$  \\
\hline
\end{tabular}
}\vspace{-0.197cm}
\end{table*}
We introduce EBMs (JEMPP~\cite{yang2021jem++}, SADAJEM~\cite{yang2023towards}, Energy Margin~\cite{liu2020energy}, MEJEM)  for dialect OOD detection. Through conducting comprehensive experiments, our goal is to showcase the effectiveness of EBMs and energy-based OOD scores, as well as to analyze their performance in dialect OOD detection tasks.
\subsection{OOD Detection Performance}
%
%
\textbf{Close-set performance.} In close-set studies, EBMs show minimal precision decrease on in-set data compared to the baseline. Notably, SADAJEM achieves higher precision (99.55\%) than softmax models (99.23\%), highlighting its effectiveness. Additionally, the SAM method notably improves the baseline precision from 99.23\% to 99.36\%.\\
\textbf{MEJEM performs best in OOD detection.}
In Table \ref{tab:ood-performance}, MEJEM emerges as the top performer in OOD detection, achieving AUROC scores of 0.925 and 0.987 for two OOD datasets, with improvements of 24.1\% and 33.6\% over the baseline, respectively. Additionally, SADAJEM also performs well, with AUROC values of 0.829 and 0.967 for two dialects, respectively, while JEMPP excels specifically in English OOD detection, surpassing all the softmax models. MEJEM  further improves results in another OOD detection metric, achieving significant reductions in FPR95, with reductions of 15.2\% and 11.8\%, outperforming the best baseline methods by 73.1\% and 93.7\%, for Latin American and English  dialects, respectively. 
%
\\
\textbf{MEJEM improves OOD detection  when learning the generative model.}  MEJEM outperforms SADAJEM and JEMPP with AUROC scores of 0.925 and 0.987, along with lower FPR95 rates of 0.223 and 0.054 for Latin American and English dialects, respectively.  
This improvement is attributed to MEJEM's integration of generative loss, energy-based margin loss, and the use of auxiliary datasets.
\\
\textbf{Energy Score vs. Softmax Score.} The analysis from Table~\ref{tab:close} and Table~\ref{tab:ood-performance} provides the following insights:
\begin{enumerate}
    \item EBMs demonstrate superior performance in OOD detection using energy scores, surpassing softmax models, despite the latter exhibiting higher precision on in-set data.
    \item The performance of softmax-based OOD detection correlates closely with their precision and increases as precision rises.
    \item Softmax models reach a ceiling in enhancing AUROC, suggesting limitations in their capability to detect OOD data.
\end{enumerate}
\textbf{Distribution of OOD detection.} 
We conduct further analysis of OOD detection distribution by comparing the distributions of in-set and out-set data in Figure~\ref{distribution}. It is observed that energy scores yield a smoother overall distribution, which is less susceptible to the peak distributions observed with softmax scores. Notably, EBMs utilizing energy scores enable a clearer differentiation between ID and OOD samples, thus facilitating more effective OOD detection.
\vspace{-0.23cm}
\subsection{Ablation Study}
\textbf{Ablation study of component in MEJEM.} 
We conduct an ablation study on the components of MEJEM to evaluate their impact on OOD detection.
In Table~\ref{tab:ablation}, including both the generative model and energy-based margin loss yields the highest AUROC values for Latin American and English dialects, using energy scores.
The integration of the generative model enhances OOD detection performance compared to scenarios where it is excluded.
Likewise, the presence of the energy-based margin loss significantly improves OOD detection performance.
\\
\textbf{Impact of parameter variations.}
We investigate the impact of parameters on both inset-classification and OOD detection performance within the MEJEM framework. Table~\ref{tab:m and step} presents the results of the ablation study conducted on the parameters $M$ and $step$. Notably, varying the margin value ($M$) and the sampling step size ($step$) affects the performance metrics, with $M=-10$ and $step=10$ yielding the highest AUROC and precision scores. Additionally, we explore the influence of loss weights $\lambda_1$ and $\lambda_2$ on dialect classification and OOD detection performance, as depicted in Table~\ref{tab:m and step}. 
\vspace{-0.13cm}
\begin{table}[htbp]
\centering
\small
\caption{Ablation Study Results for OOD Detection with SAM Optimization During Training. Values are AUROC.}
\label{tab:ablation}
 \renewcommand{\arraystretch}{1.039} 
\resizebox{6.73cm}{!}{
\begin{tabular}{cccc} \hline
Generative & Margin    &  \multicolumn{2}{c}{Softmax score / Energy score}    \\ 
Model & Loss &Latin American & English\\ \hline
$\checkmark$  & $\checkmark$  & 0.784/\textbf{0.925}  & 0.774/\textbf{0.987}  \\ 
$\checkmark$ &  &   0.752/0.829  & 0.728/0.967   \\
& $\checkmark$    & 0.711/0.875  & 0.737/0.968   \\
&     & 0.783/0.729  & 0.739/0.609    \\  \hline
\end{tabular}}
\end{table} 
\vspace{-0.53cm}

\begin{table}[htbp]
\centering\small
\caption{Effect of Parameters (Sampling Step, $\lambda_1$, Margin, $\lambda_2$). }\label{tab:m and step}
\resizebox{7.57cm}{!}{%
\begin{tabular}{ccccc}
\hline
Methods & Parameter & AUC(Latin)   & AUC(English)  & Precision    \\
\hline
WideResNet&- & 0.763 &0.733 & 99.23\%  \\ \hline
&$S$=5 &0.815  & 0.914 & 98.79\% \\
&$S$=10 &\textbf{0.881}  & 0.961 & 99.51\% \\
&$S$=15 &0.829  & \textbf{0.967} & 99.55\% \\
+Generative&$S$=20 &0.834  & 0.834 & 99.87\% \\
Loss         &$\lambda_1$=0.3  & 0.823 & 0.936 & 99.68\% \\
         &$\lambda_1$=0.6  & \textbf{0.833} & 0.945 & 99.59\% \\         &$\lambda_1$=1.0  & 0.829 & \textbf{0.967} & 99.55\% \\
         & $\lambda_1$=1.5 & 0.831 & 0.945 & 99.72\% \\ \hline
&$M$=-5 & 0.820 &0.936 & 99.08\%  \\
&$M$=-10 &\textbf{0.875} &\textbf{0.968} & 98.91\%  \\
+Margin&$M$=-15 & 0.830 &0.894 & 98.24\%  \\
Loss        &  $\lambda_2$=0.03 & 0.799 & 0.944 & 99.01\% \\
        &  $\lambda_2$=0.05 & \textbf{0.875} & \textbf{0.968} & 98.91\% \\
        &  $\lambda_2$=0.10 & 0.863 & 0.922 & 98.51\% \\ 
\hline
\end{tabular}}
\end{table}\vspace{-0.5cm}
\section{Conclusion}\label{sec:conclu}
In this study, we introduce EBMs for dialect OOD detection and propose a novel method named MEJEM. MEJEM combines generative loss with a hybrid energy-based margin loss, leveraging generative models to learn the density distribution of training data. The key idea  is to assign low energy to ID data and high energy to OOD data, creating a clear energy gap between them.
In addition, for dialect OOD detection, both energy scores and softmax scores suit different models. Despite high in-set accuracy, softmax scores hit a performance bottleneck in OOD detection. In contrast, energy scores achieve higher levels, underscoring the superior performance and potential of EBMs in addressing challenges in dialect OOD detection.
\label{section:conclu}

\newpage
\bibliographystyle{IEEEtran}
\bibliography{mybib}
\end{document}